\documentclass{article}
\usepackage{ijcai20}

\usepackage{times}

\usepackage{soul}
\usepackage{url}
\usepackage[hidelinks]{hyperref}
\usepackage[utf8]{inputenc}
\usepackage[small]{caption}
\usepackage{graphicx}
\usepackage{amsmath}
\usepackage{booktabs}
\usepackage[bitstream-charter]{mathdesign}

\usepackage[frozencache=true]{minted}
\usepackage{listings}
\usepackage{microtype}
\usepackage{tikz}
\usepackage{amsthm}
\usepackage{amsmath}
\usepackage{algorithm}
\usepackage{pgfplots}
\usepackage{subcaption}
\usepackage{inconsolata}

\newcommand{\name}{Brute}

\newcommand{\uni}{\name{}$_{\models}$}

\newcommand{\mc}[1]{$\mathcal{#1}$}

\theoremstyle{definition}
\newtheorem{definition}{Definition}
\newtheorem{example}{Example}

\definecolor{pixel 0}{HTML}{FFFFFF}
\definecolor{pixel 1}{HTML}{FF0000} 

\lstnewenvironment{myalgorithm}[1][] 
{
    \lstset{ 
        showspaces=false,               
        showstringspaces=false,         
        mathescape=true,
        numbers=left,
        escapeinside={*}{*},
        columns=flexible,
        numberstyle=\footnotesize,
        basicstyle=\footnotesize,
        keywordstyle=\footnotesize\bfseries,
        keywords={,and, return, not, def, in, if, else, for, foreach, while, }
        numbers=left,
        xleftmargin=.02\textwidth,
        #1 
    }
}
{}

\title{Learning Large Logic Programs By Going Beyond Entailment}

\author{
    Andrew Cropper$^1$
    \And
    Sebastijan Dumančić$^2$
    \affiliations
     $^1$University of Oxford\\
     $^2$KU Leuven\\
    \emails{
    andrew.cropper@cs.ox.ac.uk,
    sebastijan.dumancic@cs.kuleuven.be
    }
}

\begin{document}

\maketitle

\begin{abstract}
A major challenge in inductive logic programming (ILP) is learning large programs.
We argue that a key limitation of existing systems is that they use entailment to guide the hypothesis search.
This approach is limited because entailment is a binary decision: a hypothesis either entails an example or does not, and there is no intermediate position.
To address this limitation, we go beyond entailment and use \emph{example-dependent} loss functions to guide the search, where a hypothesis can partially cover an example.
We implement our idea in \name{}, a new ILP system which uses best-first search, guided by an example-dependent loss function, to incrementally build programs.
Our experiments on three diverse program synthesis domains (robot planning, string transformations, and ASCII art), show that \name{} can substantially outperform existing ILP systems, both in terms of predictive accuracies and learning times, and can learn programs 20 times larger than state-of-the-art systems.
\end{abstract}

\section{Introduction}

A major challenge in inductive logic programming (ILP) is learning large programs \cite{crop:iggp}.
We argue that a key limitation of existing systems is that they use entailment to guide the hypothesis search \cite{progol,tilde,aleph,ilasp,metagol}.
This approach is limited because entailment is a binary decision: a hypothesis either entails an example or does not, and there is no intermediate position.

\begin{figure}[ht]
\def\pixelsone{
  {0,1,1,0},
  {1,0,0,1},
  {1,0,0,1},
  {0,1,1,1}%
}
\def\pixelstwo{
  {1,1,1,1},
  {1,1,1,1},
  {1,1,0,1},
  {1,1,1,1}%
}

\centering
\begin{subfigure}{.5\linewidth}
\centering
\begin{tikzpicture}[scale=0.25]
  \foreach \line [count=\y] in \pixelsone {
    \foreach \pix [count=\x] in \line {
      \draw[fill=pixel \pix] (\x,-\y) rectangle +(1,1);
        \ifthenelse{\pix = 0}
        {}
        {\draw[fill=red] (\x,-\y) rectangle +(1,1);}

    }
  }
\end{tikzpicture}
\caption*{Output $o_1$}
\end{subfigure}%
\begin{subfigure}{.5\linewidth}
\centering
\begin{tikzpicture}[scale=0.25]
  \foreach \line [count=\y] in \pixelstwo {
    \foreach \pix [count=\x] in \line {
      \draw[fill=pixel \pix] (\x,-\y) rectangle +(1,1);
        \ifthenelse{\pix = 0}
        {}
        {\draw[fill=red] (\x,-\y) rectangle +(1,1);}

    }
  }
\end{tikzpicture}
\caption*{Output $o_2$}
\end{subfigure}%
\caption{Outputs $o_1$ and $o_2$ from hypotheses $h_1$ and $h_2$ respectively.}
\label{fig:intro-actions}
\end{figure}
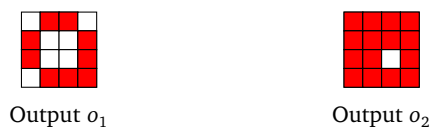

To illustrate this limitation, imagine learning image transformation programs from input/output examples.
Figure \ref{fig:intro} shows a scenario where the goal is to learn a program to transform the corner squares from red to white.
Suppose that an entailment-guided ILP system is evaluating two hypotheses $h_1$ and $h_2$ which generate the outputs $o_1$ and $o_2$ respectively shown in Figure \ref{fig:intro-actions}.
Although $o_1$ is clearly closer to the example output than $o_2$ (because only 1 pixel needs to change compared to 7 in $o_2$), the ILP system would deem the two hypotheses equal because neither entails the example, and would thus have no reason to prefer $h_1$ to $h_2$ during the search.

\def\pixelsxa{
  {1,1,1,1},
  {1,0,0,1},
  {1,0,0,1},
  {1,1,1,1}%
}
\def\pixelsya{
  {0,1,1,0},
  {1,0,0,1},
  {1,0,0,1},
  {0,1,1,0}%
}

\def\pixelsxb{
  {1,1,1,1,1},
  {1,0,0,0,1},
  {1,0,0,0,1},
  {1,0,0,0,1},
  {1,1,1,1,1}%
}
\def\pixelsyb{
  {0,1,1,1,0},
  {1,0,0,0,1},
  {1,0,0,0,1},
  {1,0,0,0,1},
  {0,1,1,1,0}%
}

\begin{figure}[ht]
\centering
\begin{subfigure}{.5\linewidth}
\centering
\begin{tikzpicture}[scale=0.25]
  \foreach \line [count=\y] in \pixelsxa {
    \foreach \pix [count=\x] in \line {
      \draw[fill=pixel \pix] (\x,-\y) rectangle +(1,1);
        \ifthenelse{\pix = 0}
        {}
        {\draw[fill=red] (\x,-\y) rectangle +(1,1);}

    }
  }
\end{tikzpicture}
\caption*{Example input}
\end{subfigure}%
\begin{subfigure}{.5\linewidth}
\centering
\begin{tikzpicture}[scale=0.25]
  \foreach \line [count=\y] in \pixelsya {
    \foreach \pix [count=\x] in \line {
      \draw[fill=pixel \pix] (\x,-\y) rectangle +(1,1);
        \ifthenelse{\pix = 0}
        {}
        {\draw[fill=red] (\x,-\y) rectangle +(1,1);}

    }
  }
\end{tikzpicture}
\caption*{Example output}
\end{subfigure}
\caption{Image transformation example.}
\label{fig:intro}
\end{figure}
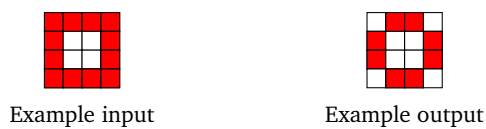

To address this limitation, we take inspiration from how humans write programs.
To paraphrase Ellis et al. (\citeyear{ellis:repl}), writing code is often a process of trial and error: write code, execute it, evaluate the output, and revise the code if necessary.
Our approach allows an ILP system to perform similarly: build a program, execute it on example input to generate output, compare the output with the expected output, and revise the program if necessary.


To allow an ILP system to evaluate a program (i.e. to compare the generated output with the expected output), we give it domain-specific knowledge.
Specifically, we use \emph{example-dependent} loss functions: loss functions that consider information about the examples, other than whether they are entailed.
For instance, in the image transformation problem, we could use Hamming distance to measure how close an output is to the desired one (how many pixels differ), which would allow an ILP system to prefer $h_1$ to $h_2$.

We claim that our approach can improve learning performance, especially when the target hypothesis is large.
To support this claim, we make the following contributions:

\begin{itemize}
\item
We describe \name{}, a new ILP system (Section \ref{sec:impl}) which combines techniques from search (best-first search) and answer set programming (ASP) \cite{asp} to learn programs with recursion and predicate invention \cite{mugg:metagold}.

\item
We evaluate \name{} on three diverse program synthesis domains: robot planning, real-world string transformations, and a new problem of drawing ASCII art (Section \ref{sec:experiments}).
Our experiments show that \name{} can substantially outperform existing ILP systems both in terms of predictive accuracies and learning times, and can learn programs 20 times larger than state-of-the-art systems.
\end{itemize}

\section{Related Work}
\label{sec:related}

The goal of program synthesis is to automatically generate computer programs from input/output examples.
The topic is considered the holy grail of AI \cite{DBLP:journals/ftpl/GulwaniPS17,DBLP:conf/snapl/SinghK17}.

Neural approaches typically need lots of training data and large training times \cite{deepcoder,DBLP:conf/nips/DevlinBSHK17,ellis:scc}.
For instance, REPL \cite{ellis:repl} combines an interpreter with AlphaGo \cite{alphago} approach to learn programs.
However, REPL takes multiple days to train on a single domain using one P100 GPU.
By contrast, \name{} can learn programs in under 60 seconds using a standard single-core computer.
Another disadvantage of neural approaches is that they often require hand-crafted neural architectures for each domain.
For instance, REPL needs a hand-crafted grammar, interpreter, and neural architecture for each domain.
By contrast, \name{} uses logic programming as a uniform representation for examples, background knowledge, hypotheses, and for itself (i.e. \name{} is written in Prolog), so can be applied to arbitrary domains.


Classical ILP systems, such as FOIL \cite{foil}, Progol \cite{progol}, TILDE \cite{tilde}, and Aleph  \cite{aleph}, cannot (in general) learn recursive programs and so often struggle on synthesis problems.
By contrast, \name{} can learn recursive programs, so can learn programs that generalise to arbitrary sized inputs.

Most ILP systems use entailment-based loss functions (also called \emph{cost} or \emph{scoring} functions) to guide the hypothesis search, often in combination with the hypothesis size \cite{foil,progol,tilde,aleph,ilasp,metagol,hexmil}.
For instance, Aleph's default loss function is $p - n$, where $p$ and $n$ are the number of positive ($p$) and negative ($n$) examples entailed by a clause.
However, as our introductory image transformation example shows, entailment-based loss functions can be uninformative because entailment is a binary decision.

Metaopt \cite{crop:metaopt} and FastLAS \cite{law:fastlas} both use \emph{domain-specific} loss functions to find optimal hypotheses, such as the most efficient program \cite{crop:metaopt}.
\name{} also uses domain-specific loss functions (specifically example-dependent), not to learn optimal programs, but to instead search efficiently.

Most authors measure the size of a logic program as either the number of literals \cite{ilasp} or clauses \cite{crop:metaho} in it.
What defines a large program is unclear.
Given $n$ examples, ILP systems can learn programs with $n$ clauses by simply memorising the examples.
ILP systems can also learn clauses with many literals when the variable depth is small \cite{progol}.
In contrast to most ILP systems, which focus on concept learning (i.e. classification), \name{} focuses on learning recursive programs that \emph{compute} something, i.e. program synthesis \cite{flener}.
In this area, Metagol \cite{metagol} is a state-of-the-art system, yet struggles to learn programs with more than 8 clauses (or approximately 24 literals) \cite{crop:iggp}.
Our experiments show that \name{} can learn programs 20 times larger than Metagol, whilst still maintaining the ability to generalise.


\name{} works in two stages: invent and search.
ILASP \cite{ilasp} and $\partial$ILP \cite{evans:dilp} also work in two stages.
Both precompute a set of clauses and then find a subset of the clauses, where ILASP uses ASP and $\partial$ILP uses a neural network.
\name{} also finds a subset of clauses, but additionally finds how to compose the clauses to build new clauses.

\section{\name{}}
\label{sec:impl}

\name{} is a new ILP system which we intentionally designed to be simple to clearly demonstrate our idea.
Given:
\begin{itemize}
\item positive ($e^+$) and negative ($e^-$) examples formed of sets of dyadic atoms $p(x_i,y_i)$ where $p$ is the target predicate symbol and $x_i$ and $y_i$ are terms denoting input and output values respectively
\item background knowledge $bk$ described as a definite logic program
\item an example-dependent loss function $\mathcal{L} : \mathcal{C} \times \mathcal{C} \rightarrow R$ where \mc{C} is the constant signature of $e^+ \cup e^- \cup bk$
\end{itemize}
\name{} searches for a hypothesis (a definite logic program) $h$ such that $\forall e \in e^+, h \cup bk \models e$ and $\forall e \in e^-, h \cup bk \not\models e$.

Brute works in two stages: \textbf{invent} and \textbf{search}.
In the invent stage, \name{} invents \emph{library predicates}, which it later uses to build a hypothesis.
In the search stage, \name{} searches to find which library predicates to add to a hypothesis and in which sequence to execute them.
Algorithm \ref{alg:brute} sketches the \name{} algorithm\footnote{\name{} is implemented in Prolog.}.
We describe the two stages in detail.

\begin{algorithm}[ht]
\begin{myalgorithm}[]
def $\text{brute}$(e$^+$,e$^-$,bk,$\mathcal{L}$):
  library = invent(bk)
  prog = search(e$^+$,e$^-$,bk,$\mathcal{L}$,library)
  return prog

def $\text{search}$(e$^+$,e$^-$,bk,$\mathcal{L}$,library):
  queue = empty_priority_queue()
  initial_spec = $\{(x,y) \;|\; p(x,y) \in e^+\}$
  initial_hypothesis = []
  initial_state = (initial_spec,initial_hypothesis)
  initial_loss = $\sum_{(x,y) \in initial\_spec} \mathcal{L}(x,y)$
  queue.push(initial_loss,initial_state)

  while not queue.empty():
    loss,state = queue.pop()
    (spec,hypothesis) = state

    if loss == 0 and consistent(hypothesis,e$^-$,bk):
      return hypothesis + induce_target(hypothesis)

    for library_predicate in library:
      new_spec = apply(library_predicate,bk,spec)
      new_hypothesis = hypothesis + library_predicate
      new_loss = $\sum_{(x,y) \in new\_spec} \mathcal{L}(x,y)$
      new_state = (new_spec,new_hypothesis)
      q.push(new_loss,new_state)
  return []
\end{myalgorithm}
\caption{\name{}}
\label{alg:brute}
\end{algorithm}

\subsection{Invent}
\label{sec:actions}

Brute takes as input background knowledge which defines \emph{primitive} predicates.
In the invent stage, \name{} uses the primitive predicates to invent \emph{library predicates}.
A \textbf{library predicate} is set of definite clauses defined by the same predicate symbol, similar to a predicate in a Prolog library, e.g. \emph{max\_list/2}.
We call the set of library predicates the \textbf{library}.

To invent library predicates, \name{} uses an ASP encoding to generate programs that compose the primitive predicates, and assigns each program a unique new predicate symbol.
Because the set of such programs is infinite, \name{} follows common convention \cite{aleph,ilasp,metagol} and restricts the maximum number of (1) clauses, (2) distinct variables in a clause, and (3) body literals in a clause.

\begin{example}
\label{ex:gen-predicates}
Given the primitive predicates \emph{right/2}, \emph{draw\_black/2}, \emph{draw\_white/2}, \emph{at\_end/1} and suitable restrictions on the number of clauses, distinct variables, and body literals, \name{} would invent the predicate:

\begin{center}
  \begin{tabular}{l}
  \emph{f1(A,B) $\leftarrow$ at\_end(A), draw\_white(A,B)}\\
  \end{tabular}
\end{center}

\noindent
\name{} would also invent recursive predicates, such as:

\begin{center}
  \begin{tabular}{l}
  \emph{f2(A,B) $\leftarrow$ at\_end(A), draw\_black(A,B)}\\
  \emph{f2(A,B) $\leftarrow$ draw\_white(A,C), right(C,D), f2(D,B)}
  \end{tabular}
\end{center}
\end{example}

\name{} uses ASP constraints to eliminate pointless predicates.
These constraints are important because they improve efficiency in the search stage by reducing the branching factor of the search tree.
Due to space restrictions we cannot detail these constraints but instead give a few examples.

\begin{example}
Given the same input as in Example \ref{ex:gen-predicates}, \name{} does not invent the following predicate (due to pruning constraints) because the literal \emph{at\_start(C)} is not connected to the rest of the clause.

\begin{center}
  \begin{tabular}{l}
  \emph{f3(A,B) $\leftarrow$ at\_start(B), at\_start(C)}\\
  \end{tabular}
\end{center}

\noindent
\name{} does not invent the following predicate because the variable $A$ is in the head but not in the body (a Datalog-like constraint):

\begin{center}
  \begin{tabular}{l}
  \emph{f4(A,B) $\leftarrow$ at\_start(B)}\\
  \end{tabular}
\end{center}

\noindent
\name{} does not invent the following predicate because it has a recursive clause without a base clause:

\begin{center}
  \begin{tabular}{l}
  \emph{f5(A,B) $\leftarrow$ right(A,C), f5(C,B)}\\
  \end{tabular}
\end{center}
\end{example}

\subsection{Search}
In the search stage, \name{} tries to build a hypothesis using the library predicates.
To do so, \name{} performs a best-first search \cite{ai:book} guided by a given loss function.

A \textbf{specification} is a set of $(x,y)$ pairs denoting input ($x$) and output ($y$) values.
Given $n$ positive examples $\{p(x_1,y_1), \dots, p(x_n,y_n)\}$, the initial specification (line 8) is $\{(x_1,y_1), \dots, (x_n,y_n)\}$.
The initial hypothesis is an empty list, denoted as $[]$ (line 9).
A \textbf{state} is a \emph{$\langle$ specification, hypothesis $\rangle$} pair.
The initial state (line 10) is a pair of the initial specification and the initial hypothesis.
The initial \textbf{loss} is the sum of the losses of the initial specification (line 11).
\name{} adds the initial loss and the initial state to a priority queue (line 12) and then performs a best-first search.

\begin{example}
Suppose we have the positive examples \emph{f(1,4) and f(7,10)}, the loss function $\mathcal{L}(x,y) = |x-y|$, and four library predicates:

\begin{center}
  \begin{tabular}{l}
  \emph{f1(A,B) $\leftarrow$ succ(A,B)}\\
  \emph{f2(A,B) $\leftarrow$ double(A,B)}\\
  \emph{f3(A,B) $\leftarrow$ double(A,C),double(C,B)}\\
  \emph{f4(A,B) $\leftarrow$ succ(A,C),succ(C,B)}\\
  \end{tabular}
\end{center}

\noindent
The initial specification $s$ is $\{(1,4), (7,10)\}$, the initial state is $\langle s, [] \rangle$ and the initial loss is $|1-4| + |7-10| = 6$.
Because the loss is not zero, \name{} searches for a better state.
\end{example}


To search for a better state, \name{} \textbf{applies} each library predicate to every pair in the current specification to generate a new specification (line 22).
For instance, applying the library predicate \emph{f1(A,B) $\leftarrow$ succ(A,B)} to the specification pair $(1,4)$ means to call \emph{f1(1,B)} to deduce a value for $B$ (i.e. 2) to form the new specification pair $(2,4)$.
Formally\footnote{We assume reader familiarity with logic programming especially the concepts of an \emph{answer substitution} and a \emph{computed answer} \cite{lloyd:book}.}:


\begin{definition}[\textbf{Application}]
Given background knowledge $bk$, a library $l$, a library predicate $p/2$, and a specification pair $(x,y)$, an \emph{application} forms a new specification pair $(z,y)$ where $z$ is the computed answer for $B$ in an SLD-refutation of $bk \cup l \cup \{\leftarrow p(x,B)\}$.
\end{definition}


\begin{example}
Applying the library predicate \emph{f4(A,B) $\leftarrow$ succ(A,C),succ(C,B)} to the specification $\{(1,4), (7,10)\}$ generates the new specification $\{(3,4), (9,10)\}$ from the computed answers of the SLD-refutations of $bk \, \cup l \cup \{\leftarrow f4(1,B_1)\}$ and $bk \cup l \cup \{\leftarrow f4(7,B_2)\}$ respectively.
\end{example}

After generating a new specification, \name{} adds the library predicate to the hypothesis (line 23), calculates the loss of the new specification (line 24), and adds the new state with the new loss to the priority queue (line 25).

\begin{example}
Continuing our running example, the state is $\langle \{(1,4), (7,10)\}, [] \rangle$ and the loss is $|1-4| + |7-10| = 6$.
\name{} can apply the library predicate \emph{f4(A,B) $\leftarrow$ succ(A,C),succ(C,B)} to generate the new state:
\[ \langle \{(3,4), (9,10)\}, [\text{\emph{f4(A,B)}} \leftarrow \text{\emph{succ(A,C), succ(C,B)}}] \rangle \]
\noindent
The loss of this new state is $|3-4| + |9-10| = 2$, so \name{} again searches for a better state.
\end{example}

The search continues until either (1) there are no more states to consider, or (2) the loss at the current state is 0 and the hypothesis does not entail any negative examples (line 18).
When the search finishes, \name{} induces a target-clause from the sequence of used library predicates and adds it to the hypothesis, which it returns (line 19).
A target clause defines the target predicate $p$ and is of the form $p(S_1,S_{m+1})\leftarrow p_1(S_1,S_2),p_2(S_2,S_3),\dots,p_m(S_m,S_{m+1})$ where each $p_i$ is the $i$th clause in the hypothesis and each $S_i$ is a variable.

\begin{example}
To finish our running example, our state is:
\[ \langle \{(3,4), (9,10)\}, [\text{\emph{f4(A,B)}} \leftarrow \text{\emph{succ(A,C), succ(C,B)}}] \rangle \]
\noindent
And our loss is $|3-4| + |9-10| = 2$.
\name{} can apply the predicate \emph{f1(A,B) $\leftarrow$ succ(A,B)} to generate the new state:
\begin{equation*}
\begin{split}
    \langle \{(4,4), (10,10)\}, [ & (\text{\emph{f4(A,B) $\leftarrow$ succ(A,C), succ(C,B))}},\\
    & (\text{\emph{f1(A,B) $\leftarrow$ succ(A,B))]}} \rangle\\
\end{split}
\end{equation*}
\noindent
The loss is now 0, so the search stops.
\name{} then induces a target clause from the sequence of library predicates and adds it to the hypothesis to form:

\begin{center}
  \begin{tabular}{l}
  \emph{f(A,B) $\leftarrow$ f4(A,C), f1(C,B)}\\
  \emph{f4(A,B) $\leftarrow$ succ(A,C), succ(C,B)}\\
  \emph{f1(A,B) $\leftarrow$ succ(A,B)}
  \end{tabular}
\end{center}

\end{example}

\section{Experiments}
\label{sec:experiments}

We claim that example-dependent loss functions can improve improve learning performance.
Our experiments therefore aim to answer the question:

\begin{description}
\item[Q1] Can example-dependent loss functions outperform entailment-based loss functions?
\end{description}

\noindent
To answer \textbf{Q1}, we compare \name{} against \uni{}, a variant which mimics an entailment-based approach using the loss function:
\[ \mathcal{L}(x,y) =
  \begin{cases}
    0 & \quad \text{if } x = y\\
    1 & \quad \text{otherwise}
  \end{cases}
\]
In other words, if the output from the hypothesis exactly matches the desired output then there is no loss; otherwise the loss is 1.

We also claim that \name{} can outperform existing ILP systems, especially when the target hypothesis is large, because it uses example-dependent loss functions to guide the search.
Our experiments therefore aim to answer the question:

\begin{description}
\item[Q2] Can \name{} outperform state-of-the-art ILP systems?
\end{description}

\noindent
To answer \textbf{Q2}, we compare \name{} against Metagol, a state-of-the-art ILP system, which can learn recursive programs.


To answer questions \textbf{Q1} and \textbf{Q2}, we consider three diverse domains: robot planning, string transformations, and drawing ASCII art.

\subsubsection{Experimental Settings}
In each experiment, we enforce a learning timeout of 60 seconds per task.
We repeat each experiment 10 times and plot 95\% confidence intervals.
We use only positive training examples.
We describe below the system settings used in the experiments.

\paragraph{\name{}}
We restrict \name{} to only invent library predicates with at most two clauses, where each clause has at most three variables and at most two body literals.

\paragraph{Metagol}
Metagol uses metarules \cite{mugg:metagold}, higher-order Horn clauses, to guide the proof search.
We provide Metagol with the \emph{ident}, \emph{precon}, \emph{postcon}, \emph{chain}, and \emph{tailrec} metarules, which are commonly used in the Metagol literature.
We also force Metagol to learn functional logic programs \cite{mugg:metabias}, which ensures that for any induced program and for any input value there is exactly one output value.
This constraint helps Metagol when learning from only positive examples because otherwise it tends to learn overly general recursive programs.

\subsection{Experiment 1 - Robot Planning}
Our first experiment is on learning robot plans, a domain often used to evaluate Metagol \cite{crop:playgol,crop:metaopt}.

\subsubsection{Materials}
A robot and a ball are in a $n^2$ grid.
An example is an atom $f(x,y)$, where $f$ is the target predicate and $x$ and $y$ are initial and final states respectively.
A state describes the positions of the robot and the ball, and whether the robot is holding the ball.
The task is to learn to transform the initial state to the final state.
We provide as background knowledge the dyadic predicates \emph{up}, \emph{down}, \emph{right}, \emph{left}, \emph{grab}, \emph{drop}, and the monadic predicates \emph{at\_top}, \emph{at\_bottom}, \emph{at\_left}, \emph{at\_right}.
\name{} uses a Manhattan distance loss function.

\subsubsection{Method}
For each $n$ in $\{2,4,6,..,10\}$, we generate a single training example for a $n^2$ world, i.e. this is a one-shot learning task.
As the grid size grows, so should the size of the target hypothesis.
We measure the percentage of tasks solved (i.e. where an induced hypothesis entails all the positive and no negative examples) and learning times.

\subsubsection{Results}
\label{sec:robores}
\begin{figure}[ht]
\centering
\begin{subfigure}{.5\linewidth}
\centering
\pgfplotsset{scaled x ticks=false}
\begin{tikzpicture}[scale=.5]
    \begin{axis}[
    xlabel=Grid size,
    ylabel=Tasks solved (\%),
    xmin=2,xmax=10,
    ymin=0,ymax=100,
    ylabel style={yshift=-5mm},
    legend style={legend pos=south west,font=\small,style={nodes={right}}}
    ]

\addplot+[mark=square*,error bars/.cd,y fixed,y dir=both,y explicit] coordinates {
(2,94.0) +- (0,4.735993637392804)
(4,100.0) +- (0,0.0)
(6,100.0) +- (0,0.0)
(8,90.0) +- (0,5.982639209833741)
(10,67.0) +- (0,9.377043475315045)
};

\addplot+[mark=oplus*,error bars/.cd,y fixed,y dir=both,y explicit] coordinates {
(2,94.0) +- (0,4.735993637392804)
(4,100.0) +- (0,0.0)
(6,89.0) +- (0,6.239688427479929)
(8,41.0) +- (0,9.808204052531282)
(10,26.0) +- (0,8.747301422885748)
};

\addplot+[black,mark=none,dashed,error bars/.cd,y fixed,y dir=both,y explicit] coordinates {
(2,100.0) +- (0,0.0)
(4,28.000000000000004) +- (0,8.953994476751133)
(6,8.0) +- (0,5.410164438810166)
(8,1.0) +- (0,1.9842169515086825)
(10,1.0) +- (0,1.9842169515086825)
};

    \legend{\name{}, \uni{}, Metagol}
    \end{axis}
  \end{tikzpicture}
\caption{Percentage of tasks solved}
\end{subfigure}%
\begin{subfigure}{.5\linewidth}
\centering
\pgfplotsset{scaled x ticks=false}
\begin{tikzpicture}[scale=.5]
    \begin{axis}[
    xlabel=Grid size,
    ylabel=Learning time (seconds),
    xmin=2,xmax=10,
    ylabel style={yshift=-5mm},
    legend style={legend pos=north west,font=\small,style={nodes={right}}}
    ]

\addplot+[mark=square*,error bars/.cd,y fixed,y dir=both,y explicit] coordinates {
(2,0.3084207272529602) +- (0,0.0037260530367211678)
(4,0.3341983914375305) +- (0,0.016143450546162355)
(6,1.3593309617042542) +- (0,0.5707293242693191)
(8,12.650155766010284) +- (0,3.812750620646456)
(10,29.386501321792604) +- (0,5.045036263732659)

};

\addplot+[mark=oplus*,error bars/.cd,y fixed,y dir=both,y explicit] coordinates {
(2,0.30472583293914796) +- (0,0.0031071388595316535)
(4,0.6338102316856384) +- (0,0.1389227271164655)
(6,13.36481141090393) +- (0,3.8342285192284)
(8,41.93340153694153) +- (0,4.8176751808781075)
(10,48.03312175273895) +- (0,4.3320512610394095)

};

\addplot+[black,mark=none,dashed,error bars/.cd,y fixed,y dir=both,y explicit] coordinates {
(2,3.9991140723228455) +- (0,2.310360122178245)
(4,46.845752825737) +- (0,4.787055561890741)
(6,55.50320454120636) +- (0,3.0520318028304176)
(8,59.43631555080414) +- (0,1.1249401600900788)
(10,59.42599387407303) +- (0,1.144704846772749)

};

    \legend{\name{}, \uni{}, Metagol}
    \end{axis}
  \end{tikzpicture}
\caption{Mean learning time}
\end{subfigure}
\caption{Robot experimental results.}
\label{fig:robot-results}
\end{figure}
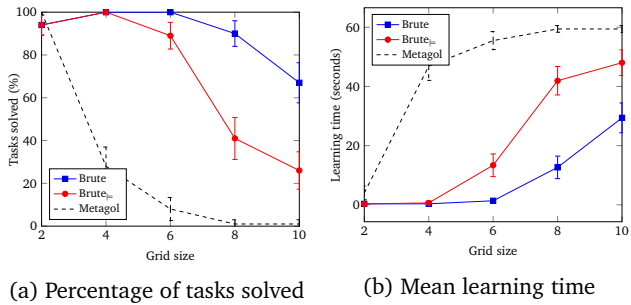

Figure \ref{fig:robot-results} shows that for a $2^2$ grid, all three systems solve 100\% of the tasks.
However, as the grid size grows, the performance of Metagol quickly degrades.
For a $6^2$ grid, Metagol only solves 8\% of the tasks.
By contrast, for a $10^2$ grid, \name{} solves 67\% of the tasks.
Figure \ref{fig:robot-results} also shows that \name{} learns programs substantially quicker than Metagol.
Figure \ref{fig:robot-results} shows that \name{} outperforms \uni{} for a $6^2$ grid or bigger, both in terms of learning times and percentage of tasks solved.
These results suggest that the answers to \textbf{Q1} and \textbf{Q2} are both yes.

Metagol struggles on larger grids because it must learn larger programs.
The biggest program learned by Metagol has 5 clauses and 15 literals.
By contrast, the biggest program learned by \name{} has 18 clauses and 69 literals.
\name{} starts to struggle on larger grids because of local optima.
For instance, suppose that the robot starts at position 1/1, the ball at position 3/1, and the goal is to move the ball to position 1/3.
In this scenario, \name{} first tries to find a program to move the robot to position 1/3 to minimise the loss function (Manhattan distance).
\name{} then explores the space around 1/3, before eventually finding the ball at position 3/1, at which point it quickly finds the target hypothesis.
Despite occasional local optima, \name{} substantially outperforms both \uni{} and Metagol.

\subsection{Experiment 2 - String Transformations}
Our second experiment is on real-world string transformations, a domain often used to evaluate program synthesis systems \cite{mugg:metabias,deepcoder,ellis:repl}.

\subsubsection{Materials}
We use 130 string transformation tasks from \cite{crop:playgol}.
Each task has 10 examples.
An example is an atom $f(x,y)$ where $f$ is the task name and $x$ and $y$ are input and output states respectively.
A state is a $(s,p)$ pair, where $s$ is the string and $p$ is a cursor pointing to a specific position in the string.
Figure \ref{fig:string-prob} shows a task with three examples, where the goal is to extract the first three letters of the month and make them uppercase.

\begin{figure}[ht]
\centering
\begin{tabular}{l|l}
\textbf{Input}                        & \textbf{Output} \\ \hline
22 July,1983 (35 years old) & JUL\\
30 October,1955 (63 years old) & OCT\\
2 November,1954 (64 years old) & NOV
\end{tabular}
\caption{Example string transformation task.}
\label{fig:string-prob}
\end{figure}

We provide as background knowledge the dyadic predicates \emph{drop}, \emph{right}, \emph{mk\_uppercase}, \emph{mk\_lowercase}, and the monadic predicates \emph{is\_letter}, \emph{is\_uppercase}, \emph{is\_space}, \emph{is\_number}, \emph{at\_start}, \emph{at\_end}.
The predicates \emph{right}, \emph{at\_start}, and \emph{at\_end} all manipulate the cursor.
The rest manipulate the string.
Brute uses a Levenshtein distance loss function.

\subsubsection{Method}
For each task and for each $n$ in $\{1,3,5,7,9\}$, we sample uniformly without replacement $n$ examples as training examples and use the other $10-n$ examples as testing examples.
We measure predictive accuracies and learning times.

\subsubsection{Results}
\begin{figure}[ht]
\centering
\begin{subfigure}{.5\linewidth}
\centering
\pgfplotsset{scaled x ticks=false}
\begin{tikzpicture}[scale=.5]
    \begin{axis}[
    xlabel=No. examples,
    ylabel=Predictive accuracy (\%),
    xmin=1,xmax=9,
    xtick={1,3,5,7,9},
    ymin=0,ymax=100,
    ylabel style={yshift=-5mm},
    legend style={legend pos=south west,font=\small,style={nodes={right}}}
    ]

\addplot+[mark=square*,error bars/.cd,y fixed,y dir=both,y explicit] coordinates {
(1,62.74461839530332) +- (0,2.5045507715630477)
(3,59.60753532182104) +- (0,1.0143763247836188)
(5,57.230499561787894) +- (0,1.3165317715184828)
(7,54.20594633792605) +- (0,1.7610519720770683)
(9,52.52100840336135) +- (0,0.6636448767512451)
};

\addplot+[mark=oplus*,error bars/.cd,y fixed,y dir=both,y explicit] coordinates {
(1,69.78962818003913) +- (0,0.7669197585919288)
(3,68.21036106750394) +- (0,1.2346079022300052)
(5,66.36722173531989) +- (0,1.377003226785981)
(7,64.43074691805656) +- (0,1.6786709522016452)
(9,59.76890756302521) +- (0,0.6468410131680484)

};

\addplot+[black,mark=none,dashed,error bars/.cd,y fixed,y dir=both,y explicit] coordinates {
(1,38.253188086943744) +- (0,0.9595947370015532)
(3,41.709320579752465) +- (0,1.0470081495450183)
(5,40.05383845348382) +- (0,0.9956684898413669)
(7,39.24013657548652) +- (0,1.5208191862628682)
(9,36.76470588235294) +- (0,1.1104902803837997)
};

    \legend{\name{}, \uni{}, Metagol}
    \end{axis}
  \end{tikzpicture}
\caption{Predictive accuracy}
\end{subfigure}%
\begin{subfigure}{.5\linewidth}
\centering
\pgfplotsset{scaled x ticks=false}
\begin{tikzpicture}[scale=.5]
    \begin{axis}[
    xlabel=No. examples,
    ylabel=Learning time (seconds),
    xmin=1,xmax=9,
    xtick={1,3,5,7,9},
    ylabel style={yshift=-5mm},
    legend style={legend pos=south east,font=\small,style={nodes={right}}}
    ]

\addplot+[mark=square*,error bars/.cd,y fixed,y dir=both,y explicit] coordinates {
(1,16.07181740265626) +- (0,0.42838286344821697)
(3,23.85917155467547) +- (0,2.9645692192591144)
(5,25.786535124595346) +- (0,1.2027381209226764)
(7,28.48981498846641) +- (0,0.7337689610627969)
(9,30.212574530564822) +- (0,0.04840322549526554)
};

\addplot+[mark=oplus*,error bars/.cd,y fixed,y dir=both,y explicit] coordinates {
(1,15.3787291600154) +- (0,0.061183296717130624)
(3,21.95063303250533) +- (0,0.2629581451416012)
(5,24.22443901208731) +- (0,0.7882833865972659)
(7,26.978363020603474) +- (0,0.23172806776486965)
(9,28.344761318426865) +- (0,0.27229683032402713)
};

\addplot+[black,mark=none,dashed,error bars/.cd,y fixed,y dir=both,y explicit] coordinates {
(1,32.70591651384647) +- (0,0.43526393725321805)
(3,34.359664363127486) +- (0,0.7610127760813795)
(5,35.9199120585735) +- (0,0.15452947708276807)
(7,36.18346859858586) +- (0,0.20406679373520828)
(9,37.63691790195612) +- (0,0.20886096862646042)
};
    \legend{\name{}, \uni{}, Metagol}
    \end{axis}
  \end{tikzpicture}
\caption{Mean learning time}
\end{subfigure}
\caption{String experimental results.}
\label{fig:string-results}
\end{figure}
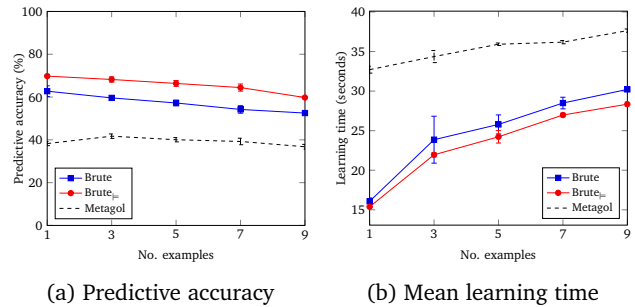

Figure \ref{fig:string-results} shows that \uni{} slightly outperforms \name{} in all cases, which contradicts the results from Experiment 1.
Figure \ref{fig:string-results} also shows that \name{} significantly outperforms Metagol in all cases, which again suggests that the answer to \textbf{Q2} is yes.

\name{} sometimes performs worse than \uni{} because of local optima.
For instance, when trying to learn a program that takes a string and returns the first letter made uppercase, e.g. \emph{``james''} $\mapsto$ \emph{``J''}, \name{} first uses an invented recursive predicate to delete all but the last character from the input to minimise the loss (edit distance), e.g. \emph{``james''} $\mapsto$ \emph{``s''} (where the loss is only 1).
\name{} then searches in this region of the search space, but is unable to find the target hypothesis in the allocated time (60 seconds).


\name{} typically outperforms \uni{} and Metagol on tasks that require larger programs.
For instance, consider trying to learn a program to extract the number of logical inferences per second (Lips) from the output of \emph{time/1} in Prolog, e.g. \emph{``16,079 inferences, 0.003 CPU in 0.003 seconds (95\% CPU, 5842660 Lips)''} $\mapsto$ \emph{``5842660''}.
For this task, \name{} learns the general program shown in Figure \ref{fig:string-prog}, which contains four invented recursive predicates.
Figure \ref{fig:execution} shows the execution trace of this program on the aforementioned example.

\begin{figure}[ht]
\centering
\begin{minipage}{.7\linewidth}
\begin{minted}[frame=single,fontsize=\footnotesize]{prolog}
f(A,B):-f0(A,C),f1(C,D),f0(D,E),
        f2(E,F),f3(F,G),f2(G,B).
f0(A,B):-is_uppercase(A),drop(A,B).
f0(A,B):-drop(A,C),f0(C,B).
f1(A,B):-is_number(A),drop(A,B).
f1(A,B):-drop(A,C),f1(C,B).
f2(A,B):-is_space(A),drop(A,B).
f2(A,B):-drop(A,C),f2(C,B).
f3(A,B):-at_end(A),drop(A,B).
f3(A,B):-right(A,C),f3(C,B).
\end{minted}
\end{minipage}
\caption{Program learned by \name{} for task \emph{p49}.}
\label{fig:string-prog}
\end{figure}

\begin{figure}
\centering
\footnotesize
  \begin{tabular}{c}
    \emph{``16,079 inferences, 0.003 CPU in 0.003 seconds (95\%}\\
    \emph{CPU, 5842660 Lips)''}\\
    $\downarrow$\;\emph{f0}\\
    \emph{``PU in 0.003 seconds (95\% CPU, 5842660 Lips)''}\\
    $\downarrow$\;\emph{f1}\\
    \emph{``.003 seconds (95\% CPU, 5842660 Lips)''}\\
    $\downarrow$\;\emph{f0}\\
    \emph{``PU, 5842660 Lips)''}\\
    $\downarrow$\;\emph{f2}\\
    \emph{``5842660 Lips)''}\\
    $\downarrow$\;\emph{f3}\\
    \emph{``5842660 Lips''}\\
    $\downarrow$ \emph{f2}\\
    \emph{``5842660''}\\
  \end{tabular}
\caption{Execution trace of the program from Figure \ref{fig:string-prog} on the example \emph{``16,079 inferences, 0.003 CPU in 0.003 seconds (95\% CPU, 5842660 Lips)''} $\mapsto$ \emph{``5842660''}.}
\label{fig:execution}
\end{figure}

\subsection{Experiment 3 - ASCII Art}
Our third experiment is on a new problem of learning to draw ASCII art.

\subsubsection{Materials}
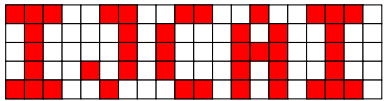
\begin{figure}[ht]

\def\ijcaipixels{
{1, 1, 1, 0, 0, 1, 1, 0, 0, 1, 1, 0, 0, 1, 0, 0, 1, 1, 1, 0},
{0, 1, 0, 0, 0, 0, 1, 0, 1, 0, 0, 0, 1, 0, 1, 0, 0, 1, 0, 0},
{0, 1, 0, 0, 0, 0, 1, 0, 1, 0, 0, 0, 1, 1, 1, 0, 0, 1, 0, 0},
{0, 1, 0, 0, 1, 0, 1, 0, 1, 0, 0, 0, 1, 0, 1, 0, 0, 1, 0, 0},
{1, 1, 1, 0, 0, 1, 0, 0, 0, 1, 1, 0, 1, 0, 1, 0, 1, 1, 1, 0}%
}
\centering
\begin{tikzpicture}[scale=0.25]
  \foreach \line [count=\y] in \ijcaipixels {
    \foreach \pix [count=\x] in \line {
      \draw[fill=pixel \pix] (\x,-\y) rectangle +(1,1);
        \ifthenelse{\pix = 0}
        {}
        {\draw[fill=red] (\x,-\y) rectangle +(1,1);}

    }
  }
\end{tikzpicture}
\caption{Example ASCII image.}
\label{fig:pixel-example}
\end{figure}

An image is the pixel representation of an ASCII string according to the text2art\footnote{https://pypi.org/project/text2art/} library with the font \emph{3x5}.
Figure \ref{fig:pixel-example} shows an example image for the string \emph{``IJCAI''}.
An example is an atom $f(x,y)$, where $f$ is the target predicate and $x$ and $y$ are input and output states respectively.
A state is a $(i,p)$ pair, where $i$ is the image, represented as a list, and $p$ is a cursor pointing to a specific pixel in the image.
We provide as background knowledge the dyadic predicates \emph{up}, \emph{down}, \emph{right}, \emph{left}, \emph{draw1}, \emph{draw0}, and the monadic predicates \emph{at\_top}, \emph{at\_bottom}, \emph{at\_left}, \emph{at\_right}.
The predicates \emph{draw0} and \emph{draw1} manipulate the image.
The rest manipulate the cursor.
Brute uses a Hamming distance loss function.

\subsubsection{Method}
For each $n$ in $\{1,2,\dots,5\}$, we sample uniformly at random with replacement an ASCII string of length $n$.
As the string grows, so should the size of the target hypothesis.
We use the text2art library to transform a string to a pixel representation which forms our output image.
We use the empty image as the input image.
We measure the percentage of tasks solved and learning times.

\subsubsection{Results}
\begin{figure}[ht]
\centering
\begin{subfigure}{.5\linewidth}
\centering
\pgfplotsset{scaled x ticks=false}
\begin{tikzpicture}[scale=.5]
    \begin{axis}[
    xlabel=No. symbols,
    ylabel=Tasks solved (\%),
    xmin=1,xmax=5,
    ymin=0,ymax=100,
    ylabel style={yshift=-5mm},
    legend style={legend pos=south west,style={nodes={right}}}
    ]

\addplot+[mark=square*,error bars/.cd,y fixed,y dir=both,y explicit] coordinates {
(1,100.0) +- (0,0.0)
(2,97.5) +- (0,3.4963112285997453)
(3,86.25) +- (0,7.712029350344452)
(4,56.25) +- (0,11.10933504622018)
(5,25.0) +- (0,9.697022622843752)
};

\addplot+[mark=oplus*,error bars/.cd,y fixed,y dir=both,y explicit] coordinates {
(1,0.0) +- (0,0.0)
(2,0.0) +- (0,0.0)
(3,0.0) +- (0,0.0)
(4,0.0) +- (0,0.0)
(5,0.0) +- (0,0.0)
};

\addplot+[black,mark=none,dashed,error bars/.cd,y fixed,y dir=both,y explicit] coordinates {
(1,0.0) +- (0,0.0)
(2,0.0) +- (0,0.0)
(3,0.0) +- (0,0.0)
(4,0.0) +- (0,0.0)
(5,0.0) +- (0,0.0)
};

    \legend{\name{}, \uni{}, Metagol}
    \end{axis}
  \end{tikzpicture}
\caption{Percentage of tasks solved}
\end{subfigure}%
\begin{subfigure}{.5\linewidth}
\centering
\pgfplotsset{scaled x ticks=false}
\begin{tikzpicture}[scale=.5]
    \begin{axis}[
    xlabel=No. symbols,
    ylabel=Learning time (seconds),
    xmin=1,
    xmax=5,
    ylabel style={yshift=-5mm},
    legend style={legend pos=south east,font=\small,style={nodes={right}}}
    ]

\addplot+[mark=square*,error bars/.cd,y fixed,y dir=both,y explicit] coordinates {
(1,0.7649444818496705) +- (0,0.11360629595353613)
(2,5.83881686925888) +- (0,2.224339498936394)
(3,19.517527732253075) +- (0,4.229776025063335)
(4,40.6979064643383) +- (0,4.752354712274157)
(5,51.50643539428711) +- (0,3.6384789921535883)
};

\addplot+[mark=oplus*,error bars/.cd,y fixed,y dir=both,y explicit] coordinates {
(1,60.004338684678075) +- (0,0.0006261655432572955)
(2,60.00367141962052) +- (0,0.00023045151134917216)
(3,60.003883531689645) +- (0,0.00042028492346102823)
(4,60.00396424531937) +- (0,0.00039595644169423364)
(5,60.00342857539654) +- (0,0.00023514292447700307)

};

\addplot+[black,mark=none,dashed,error bars/.cd,y fixed,y dir=both,y explicit] coordinates {
(1,60.003372275829314) +- (0,0.00018050080698841652)
(2,60.00362253189087) +- (0,0.00015502881649036667)
(3,60.00344600975514) +- (0,0.0001565041215345361)
(4,60.003095442056654) +- (0,0.0001914790138414176)
(5,60.002713456749916) +- (0,0.00018484911747116737)
};

    \legend{\name{}, \uni{}, Metagol}
    \end{axis}
  \end{tikzpicture}
\caption{Mean learning time}
\end{subfigure}
\caption{ASCII art experimental results.}
\label{fig:pixel-results}
\end{figure}
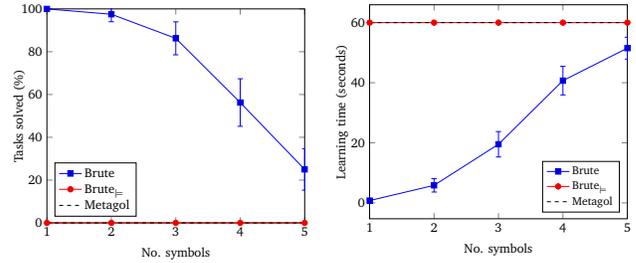

Figure \ref{fig:pixel-results} shows that Metagol and \uni{} cannot learn any solutions.
By contrast, \name{} learns programs for 86\% of the images with 3 characters, and still manages to learn programs for 25\% of the images with 5 characters.
The largest program learned by \name{} has 79 clauses and 294 literals.
These results again suggest that the answer to questions \textbf{Q1} and \textbf{Q2} is yes.

\section{Conclusions and Limitations}
A major challenge in ILP (and program synthesis in general) is learning large programs.
To tackle this problem, we have proposed an approach inspired by how humans write programs.
In our approach, an ILP system builds a program, executes it on some example input to generate output, compares the output with the expected output, and revises the program if needed.
To evaluate a hypothesis, we use \emph{example-dependent} loss functions.
We implemented our idea in \name{}, a new ILP which first invents a library of predicates, including recursive predicates, and then performs a best-first search informed by a given loss function to build a hypothesis using the library predicates.

Our experiments on three diverse program synthesis domains (robot planning, real-world string transformations, and a new problem of drawing ASCII art), show that (1) example-dependent loss functions can outperform entailment-based loss functions, and (2) \name{} can outperform Metagol, a state-of-the-art ILP system.
In our experiments, given only 60 seconds to learn a program, the largest program learned by Metagol had 5 clauses and 15 literals.
By contrast, the largest program learned by \name{} had 79 clauses and 294 literals.



\subsection*{Limitations and Future Work}

\paragraph{Generality}
In contrast to classical ILP systems, which focus on concept learning (i.e. classification), \name{} focuses on learning recursive programs from input/output examples.
\name{} cannot therefore currently solve some classical ILP problems, such as Mutagenesis \cite{srinivasan1994mutagenesis}, because the examples in these domains are not dyadic.
To address this limitation, we could represent classification tasks as program synthesis tasks, where the output is the label.

\paragraph{Search}
To clearly demonstrate our idea, we intentionally designed \name{} to be simple in two ways (1) \emph{brute}-force inventing library predicates, and (2) using simple best-first search.
As our experiments show, our intentionally simple approach can drastically outperform existing systems.
To further improve performance, we want to (1) dynamically invent library predicates during the search to reduce the branching factor, and (2) use better search techniques, such as A* or iterative budgeted exponential search \cite{ibes}.

\paragraph{Loss functions}
\name{} uses example-dependent (domain-specific) loss functions to guide the search, which our experiments show are important for high performance.
By contrast, most ILP systems use general entailment-based loss functions.
An important direction for future work is to bridge the gap between the two approaches.
An exciting idea is to \emph{learn} suitable loss functions for a given problem through \emph{meta-learning} \cite{thrun:ltl}.

\subsubsection*{Summary}
To conclude, we think that \name{} is an important contribution to ILP and program synthesis, and that the ideas introduced in this paper open up new and exciting research opportunities for learning large programs.

{
\bibliographystyle{named}
\bibliography{ourbib15}
}
\end{document}